\documentclass[11pt]{article}

\usepackage[preprint]{acl}
\usepackage{times}
\usepackage{latexsym}
\usepackage{hyperref}
\usepackage{booktabs}
\usepackage{amsmath}
\usepackage{amsfonts}
\usepackage{multirow}
\usepackage{cleveref}
\usepackage[inline]{enumitem}
\usepackage{tabularx}
\usepackage{bbm}

\newcommand{\Judge}{\text{PrivacyConcernJudge}}

\newcommand{\PRA}{\mathrm{PrivacyReasoner}}
\usepackage{url}
\usepackage[T1]{fontenc}

\usepackage[utf8]{inputenc}

\usepackage{microtype}

\usepackage{inconsolata}

\usepackage{graphicx}

\title{PrivacyReasoner: Can LLM Emulate a Human-like Privacy Mind?}

\author{Yiwen Tu \and Xuan Liu \and Lianhui Qin \and Haojian Jin \\
        University of California, San Diego}

\begin{document}
\maketitle
\begin{abstract}

Prior work on LLM-based privacy focuses on norm judgment over synthetic vignettes, rather than how people think about a specific data practice and formulate their opinions.
We address this gap by designing \(\PRA\), an agent architecture grounded in three key ideas: (1) LLMs can detect subtle privacy cues in natural language and role-play human characteristics; (2) a user's ``privacy mind'' can be reconstructed from their real-world online comment history, distilling experiences, personality, and cultural orientations; and (3) a contextual filter can dynamically activate relevant privacy beliefs based on the contexts in a scenario.
We evaluate \(\PRA\) on real-world privacy discussions from Hacker News, using an LLM-as-a-Judge evaluator calibrated against an established privacy concern taxonomy to quantify reasoning faithfulness. \(\PRA\) significantly outperforms baselines in predicting individual privacy concerns and generalizes across different domains, such as AI, e-commerce, and healthcare.

\end{abstract}

\section{Introduction}
\begin{figure*}
    \centering
    \includegraphics[width=\linewidth]{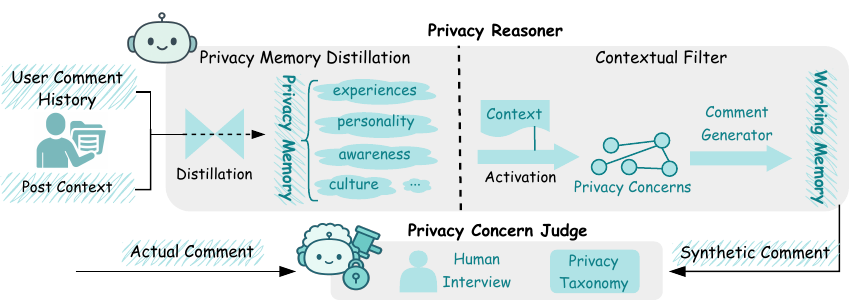}
\caption{Overview of the  \(\PRA\) workflow, an agent designed to mimic how individuals react to privacy-related events. These events are simulated using discussions from \emph{HackerNews}, a technical forum. The agent constructs its \emph{privacy memory} from two components: (1) the user’s historical comments, which serve as the raw substrate for building the agent’s profile, and (2) the post context, which includes metadata such as the post title, body text, and post initiator ID. Conditioned on this memory, the agent employs a contextual filter to selectively activate relevant beliefs and generate a synthetic comment simulating the user’s likely response. Both the synthetic and real comments are then evaluated by a \emph{Privacy Concern Judge}, which leverages an established privacy concern taxonomy and human-annotated interview data to assess concern-level alignment.}
    \label{fig:workflow}
\end{figure*}

Public reactions to privacy-related events are often rapid, polarized, and difficult to predict, yet they can shape policy, product adoption, and institutional trust. Apple's CSAM scanning feature, for instance, was technically constrained and privacy-aware by design, but triggered widespread backlash over fears of normalized surveillance, leading Apple to suspend the rollout~\cite{ApplesDe68:online}. This was not a failure of engineering, but a failure to anticipate how users would reason about the data practice~\cite{whitten1999johnny}. If an AI agent could emulate how individual users think about privacy, development teams could simulate such reactions before deployment, policy makers could stress-test regulations against realistic privacy personas, and researchers could study individual variation without costly human studies~\cite{aher2023using}.

Existing LLM-based approaches to privacy, however, do not address this challenge. Systems such as PrivacyLens~\cite{shao2024privacylens} and ConfAIde~\cite{confaide2023} frame the problem as norm judgment: given predefined contextual factors (data type, recipient, purpose), the model decides whether an information flow is appropriate. These approaches rely on synthetic vignettes constructed from Contextual Integrity parameters~\cite{nissenbaum2004privacy} and evaluate whether LLMs can match human judgments on these structured scenarios~\cite{apthorpe2018discovering}. While useful for benchmarking norm awareness, this paradigm has two key limitations. First, it reduces privacy to binary judgments of appropriateness, missing the rich reasoning process by which people weigh trade-offs, draw on past experiences, and formulate opinions. Second, it operates on artificial scenarios rather than nuanced real-world data practices, leaving open the question of whether LLMs can model how real users actually think and respond to privacy events. Preliminary evidence~\citep{10.1145/3613904.3642500} suggests that individuals vary in their sensitivity to different contextual factors, further motivating the need for individualized modeling.

We introduce \(\PRA\), an agent design that moves from norm judgment to privacy reasoning. \(\PRA\) is grounded in three key ideas. First, LLMs possess perceptual skills that detect subtle cues in implicit language and strong role-playing capabilities that link observed behaviors to human characteristics, enabling them to serve as the cognitive substrate for simulating individual privacy reasoning~\citep{aher2023using, park2023generative}. Second, a user's ``privacy mind'' can be reconstructed from their real-world comment history by distilling experiences, personality traits, awareness levels, and cultural orientations into a structured representation, grounded in the Antecedents--Privacy Concerns--Outcomes (APCO) framework~\citep{smith2011information}. Third, a contextual filter can dynamically activate relevant privacy beliefs based on the contexts in a scenario, operationalizing bounded rationality from Working Memory Theory~\citep{baddeley2020working}: not all beliefs are active at once, but context determines which ones surface.

We evaluate \(\PRA\) on real-world privacy discussions from \emph{HackerNews}~\citep{hackernews}, a technical forum where users debate surveillance, data governance, and algorithmic decision-making. To quantify reasoning faithfulness, we implement a Privacy Concern Judge, an LLM-as-a-Judge model calibrated against an established privacy concern taxonomy~\citep{10.1145/3613904.3642500}. \(\PRA\) achieves a macro F1-score of 0.473, outperforming the strongest baseline (RAG, 0.449) and substantially exceeding naive prompting (0.335) and privacy persona (0.403). Ablation studies confirm the importance of both APCO-structured distillation and contextual filtering, with the full architecture improving F1 by 41\% over naive baselines. The architecture also generalizes across domains, maintaining F1 scores of 0.410 and 0.407 on e-commerce and healthcare when trained solely on AI-domain histories.

We make the following contributions:

\begin{enumerate}[leftmargin=1em, itemsep=0pt, topsep=0pt]
    \item We reframe LLM-based privacy from norm judgment to privacy reasoning, and identify three key ideas---LLM perception, privacy mind reconstruction, and contextual activation---that guide agent architecture design for modeling individual users.

    \item We propose \(\PRA\), an agent architecture that reconstructs user-specific ``privacy minds'' from real-world comment histories and dynamically activates context-relevant beliefs, achieving substantial gains over baseline agents and generalizing across domains and individuals.

\end{enumerate}

\section{Related Work}
Our work draws on three threads: LLM-based simulation of human behavior, LLM approaches to privacy, and social media data mining for user-level analysis.

\subsection{LLM-based Simulation}
Recent work has shown that LLM-based agents can model human reasoning with population-level validity~\cite{ExplorLLMcollabration_ACL2024,Role-playing_EMNLP2024,CogBias_iclr2025,Roleplaying_EMNLP2023,park2023generativeagentsinteractivesimulacra,shaikh2025creatinggeneralusermodels, OpinionDynamics_NAACL2024, ControlCogBia_preprint2025}. For example, ~\citet{park2023generativeagentsinteractivesimulacra} introduced “generative agents,” which augment LLMs with long-term memory, reflection, and iterative planning to produce coherent daily routines and emergent social behaviors in a simulated small-town setting. Similarly, ~\citet{shaikh2025creatinggeneralusermodels} proposed General User Models, which infer confidence-weighted propositions about users from multimodal computer interactions, moving beyond self-reports to passively learned personal models. While prior research has primarily focused on collective behaviors, our work shifts focus to modeling individual privacy reasoning. Harnessing real-world datasets, we aim to simulate user-specific privacy concerns. 

\subsection{LLM for Privacy} 
Prior work treats contextual privacy chiefly as information-flow appropriateness~\cite{ICLR2024_ConfAIde, NIPS2024_PrivacyLens, li2025123checkenhancingcontextual, privacyNAACL_2025, yi2025privacy}. \citet{ICLR2024_ConfAIde} benchmarks Contextual Integrity–based reasoning and reveals frequent inappropriate disclosures by state-of-the-art large language models. \citet{NIPS2024_PrivacyLens} constructs seeds, vignettes, and trajectories to probe both question answering and agent actions, quantifying leakage in realistic tasks. \citet{li2025123checkenhancingcontextual} reduces leakage via a multi-agent checking pipeline, and \citet{privacyNAACL_2025} operationalizes Contextual Integrity with comprehensive checklists grounded in regulations such as the Health Insurance Portability and Accountability Act. \citet{yi2025privacy} studies privacy reasoning in ambiguous contexts. In contrast, we adopt a human-centered, individualized perspective: simulating aspects of human cognition to predict and explain user-specific privacy concerns, rather than merely enforcing generic disclosure policies.

\subsection{Social Media Data Mining}
Social media has been widely mined with sentiment analysis and topic modeling to surface aggregate attitudes and community norms~\cite{KDD11,KDD10,EMNLP2022-tweetnlp,aacl2020}. However, user-level analyses face persistent data sparsity—individual users often have few posts—motivating strong general user models with lightweight customization or few-shot conditioning to capture user-specific preferences while preserving robustness. Building on these observations, we emphasize individualized, context-sensitive analyses grounded in platform discourse under sparse user histories.
\section{Methodology}
Existing work on privacy reasoning with LLM-based agents has predominantly focused on \textit{information access control}---whether sensitive data can be accessed, inferred, or exposed during training or deployment. While important, this perspective captures only a narrow aspect of privacy. Human reasoning extends beyond access control to include deeper \textit{beliefs}, \textit{values}, and \textit{trade-offs} that shape how individuals interpret and respond to privacy risks. Accordingly, rather than modeling LLMs solely as access-control decision makers, we frame privacy reasoning as a \textbf{generative prediction problem}: given a concrete privacy context (e.g., a forum post), how would an agent respond in a way that faithfully reflects an individual's privacy preferences and concerns?

Our agent architecture directly operationalizes the three key ideas introduced in Section~1. 
The first idea---that LLMs can detect implicit cues and role-play human characteristics---provides the cognitive foundation. Prior work~\citep{10.1145/3613904.3642500} showed that incorporating explicit contextual labels improves privacy prediction; we go one step further by leveraging LLMs' perceptual skills to detect subtle cues directly from implicit language, removing the need for predefined labels. The second and third ideas map onto the two stages of the architecture described below.

\subsection{Privacy Reasoner}
We design a \textbf{two-stage agent architecture} that decouples the estimation of stable, person-specific privacy factors from their context-dependent activation at inference time.

\paragraph{Stage 1: Privacy Memory Distillation}
Our second idea posits that a user's ``privacy mind'' can be reconstructed from their comment history. We operationalize this by grounding the offline representation in the \textbf{Antecedents--Privacy Concerns--Outcomes (APCO)} framework~\citep{smith2011information}, which models privacy reasoning as a causal process in which privacy concerns arise from a set of relatively stable antecedent factors, including 
prior privacy experiences, 
privacy awareness, 
personality traits,
demographic characteristics,
cultural background, and subsequently influence observable behavioral outcomes. 

We prompt the LLM, leveraging its perceptual skills to detect subtle cues in implicit language, to extract and summarize atomic privacy-related factors from the user's historical data along each of the five APCO antecedent dimensions. For each dimension, the model produces a concise natural-language descriptor that captures recurrent patterns and stable tendencies evidenced across the user's prior statements, rather than transient opinions tied to a single context. These descriptors form a persistent \emph{privacy memory} that encodes the individual's latent privacy predispositions---their reconstructed ``privacy mind.'' 
An example distilled atomic privacy memory is shown below:
\begin{quote}
\footnotesize
\ttfamily
[Privacy Experiences] The user is critical of restricted internet offerings being framed as acceptable substitutes for unrestricted access.
\end{quote}
Notably, the memory distillation step is performed offline and amortized across multiple predictions, which reduces per-instance cost at inference time.
\paragraph{Stage 2: Contextual Filtering}
Our third idea holds that a contextual filter can dynamically activate the privacy orientations most relevant to each new scenario. We implement this by drawing on \textbf{Working Memory Theory}~\citep{baddeley2020working}, which holds that human reasoning is constrained by limited working memory and can engage with only a small subset of beliefs at once. We also draw on \textbf{Contextual Integrity}~\citep{nissenbaum2004privacy}, which conceptualizes privacy as adherence to context-specific norms governing information flows. Given target context, the subset of privacy memory most salient for the current situation is selected dynamically. In practice, we prompt the LLM jointly with the set of distilled privacy memories and the current context, and it outputs a filtered, ranked set of contextually relevant privacy orientations. This selective activation determines \emph{which} orientations surface for a given scenario, enabling context-sensitive reasoning rather than exhaustive recall. 
\section{Experiment Setup}

This section describes our experimental setup, including the dataset, evaluation method, the LLM-as-a-PrivacyConcernJudge, and baselines.

\subsection{Dataset}\label{sec:dataset}
\emph{HackerNews} dataset~\citep{hackernews} is a large-scale, publicly available, community-moderated social news platform. For large-scale analysis, we utilize the \texttt{bigquery-public-data.hacker\_news} dataset on Google BigQuery, which mirrors HN content from its launch in 2006 to the present. Since our objective is to model user reactions rather than conversational dynamics, we restrict our analysis to \emph{first-level comments}—direct replies to the original post—while excluding follow-up responses. The “training set” for each user is constructed by sampling that user’s past comments, with its size determined by the privacy memory parameter.

\subsection{Evaluation Method}\label{sec:evaluation}

We evaluate the reasoning process by modeling the privacy concerns expressed in generated text. Rather than treating privacy reasoning as a black box with only behavioral outcomes, we use privacy concerns as an interpretable intermediate construct that bridges data practices and downstream judgments. This approach follows prior work showing that free-form generation reveals richer reasoning patterns than structured outputs~\citep{tam-etal-2024-speak, schall-de-melo-2025-hidden}, and that privacy concerns can be systematically extracted from natural language to assess reasoning quality~\citep{10.1145/3613904.3642500}. Concretely, \(\PRA\) generates full comments rather than directly predicting privacy concern labels. This design is essential for two reasons. First, natural language comments capture richer semantic, contextual, and stylistic information than discrete labels, enabling analysis of not only \emph{what} concerns are expressed but also \emph{how} they are articulated. Second, since gold labels are derived from real comments via an LLM-as-a-Judge classifier, generating synthetic comments and processing them through the same pipeline ensures a shared interpretive filter, reducing label mismatch and enabling fairer evaluation. Both the real and generated comments are then classified using \(\Judge\) (Section~\ref{sec:judge}) against a taxonomy of 14 privacy concerns~\citep{10.1145/3613904.3642500} (\Cref{tab:privacy_concerns}), yielding multi-label predictions that we compare using three standard metrics:

\begin{itemize}[leftmargin=1em, itemsep=0pt, topsep=2pt]
    \item \textbf{Accuracy}: the overall correctness of predicted concern labels.
    \item \textbf{Recall}: how comprehensively the method identifies relevant concerns.
    \item \textbf{Macro-F1 score}: balances precision and recall across all concern categories, serving as our primary evaluation metric.
\end{itemize}

\begin{table}[t]
\scriptsize
\caption{Taxonomy of 14 Privacy Concerns from~\citet{10.1145/3613904.3642500}.}
\label{tab:privacy_concerns}
\begin{tabular}{p{0.2\textwidth} p{0.22\textwidth} }
\midrule
Lack of trust for algorithms & No control \\
Insufficient anonymization & Lack of respect for autonomy \\
Bias or discrimination & Data Leakage \\
Deception & Lack of informed consent \\
Invasive monitoring & Data commodification \\
Lack of an alternative choice & High risks \\
Unexpectation & Lack of protection for the vulnerable \\
\bottomrule
\end{tabular}
\normalsize
\end{table}

\subsection{LLM-as-a-PrivacyConcernJudge}\label{sec:judge}
Since privacy concerns are difficult to quantify, we use a well-established taxonomy from~\citep{10.1145/3613904.3642500} (\Cref{tab:privacy_concerns}) to formulate the emulation process as a multi-label classification problem. 
We construct a LLM-as-a-judge \(\Judge\) using gpt-4o~\citep{openai2024gpt4ocard}. 
Specifically, we construct a set of few-shot examples sampled from an expert-labeled dataset~\citep{10.1145/3613904.3642500} to help the model better grasp the meaning of each concern category. To assess the reliability of the LLM-based PrivacyConcernJudge, we conduct a small-scale human validation study. We randomly sample 100 comments from the HackerNews dataset and ask two human annotators to independently label the expressed privacy concerns using the same 14-category taxonomy. We then compare the LLM judge’s predictions against the human annotations. The PrivacyConcernJudge achieves substantial agreement with human labels, with a macro-averaged Cohen’s \(\kappa\) being 0.812, which suggests  "Almost perfect agreement"~\citep{Cohenska88:online}. 

\subsection{Baselines}
We introduce three baselines to understand the performance of  \(\PRA\).

\paragraph{Privacy Persona} This agent architecture is grounded in \textbf{privacy taxonomy theory}, which posits that individuals can be categorized into distinct \emph{privacy personas} based on their attitudes, knowledge, and behaviors. These personas capture stable, coarse-grained patterns in how different groups perceive and respond to privacy risks. Concretely, Privacy Persona agent embeds \textbf{attitudinal priors} into generation by conditioning synthetic comment production on an inferred \emph{privacy persona}. We consider the taxonomy proposed by~\citet{10.1145/2858036.2858214}, which distinguishes five personas: \emph{Fundamentalists}, \emph{Lazy Experts}, \emph{Self-Educated Technicians}, \emph{Amateurs}, and the \emph{Marginally Concerned}. For a given user, we infer a persona from their historical comments using an LLM-based persona classifier. The agent then conditions its reasoning and comment generation on both the post context and the inferred persona. A sample persona-conditioned prompt is as follows:
\begin{quote}
\footnotesize
\ttfamily
You are a \emph{Self-Educated Technician} when it comes to privacy. You are highly knowledgeable about technology, skeptical of corporate data practices, and prefer to retain fine-grained control over your information. Given the following post context and user history, write a comment that reflects your reasoning, concerns, and tone.
\end{quote}

\paragraph{Retrieval Augmented Generation}
To compare with current state-of-the-art approaches, we implement a Retrieval Augmented Generation(RAG)-based baseline. For each user, during offline stage, we index their historical privacy-related comments using sentence-level embeddings. We use RoBERTa~\citep{liu2019robertarobustlyoptimizedbert}. At inference time, given a target context, the agent retrieves the top-\(k\) historical comments with the highest semantic similarity to the context. These retrieved comments, first summarized by an LLM agent, are then concatenated with the current context and provided to the LLM as additional conditioning information for response generation.

\paragraph{Naive Baseline} We also include a \textbf{Naive Baseline} that does not incorporate user-specific reasoning. It receives only a simple instruction prompt, for example:
\begin{quote}
\footnotesize
\texttt{Given the following discussion post, write a comment that reflects your privacy concerns.}
\end{quote}
\section{Experimental Results}
We evaluate our approach in three stages. First, we compare overall performance against baselines. Next, we conduct an ablation study on \(\PRA\). Finally, we analyze generalization under three settings: \emph{User Transfer}, which varies the amount of available user history, \emph{Domain Transfer}, which tests whether profiles learned in one domain transfer to others, \emph{Dataset Transfer}, which tests whether our method can generalize to different dataset. We also present results demonstrating generalization across different LLM base model in~\Cref{sec:base_model}.

\subsection{Performance Comparison}~\label{sec:main}
Table~\ref{tab:main_results} reports the main performance comparison on HackerNews Dataset. \(\PRA\) consistently outperforms all baselines across metrics. The naive baseline, lacking user-specific information, approximates random guessing. Privacy Persona methods capture coarse, static privacy attitudes but fail to model context-dependent reasoning, yielding limited improvements. RAG-based methods leverage retrieved user history, but their gains are constrained by shallow retrieval and the absence of structured reasoning over retrieved evidence. Overall, these results indicate that accurate privacy reasoning requires structured, theory-grounded representations that capture both the content and dynamics of individual beliefs, rather than relying on coarse personas or unstructured retrieval.
\subsection{Ablation Study}~\label{sec:type2}

\paragraph{APCO vs.\ Unstructured Distillation.}
We evaluate whether imposing APCO-based structure during privacy memory distillation is necessary. We replace the offline stage with an unstructured variant that extracts concise privacy statements from user history without categorization or semantic role assignment. As shown in Table~\ref{tab:ablation_results}, the APCO-structured model consistently outperforms this baseline across metrics. This indicates that organizing atomic beliefs along theory-grounded antecedent dimensions yields more stable, discriminative representations, enabling more reliable contextual activation and downstream reasoning. In contrast, unstructured distillation conflates heterogeneous factors, leading to noisier relevance estimation and weaker robustness under domain shift.

\paragraph{Contextual Filter vs.\ Summary.}
We assess the importance of contextual filtering by comparing against a summary-based variant of \(\PRA\). While both share the same distillation, the summary variant compresses antecedent factors into a single representation instead of preserving structured, fine-grained beliefs. Results in Table~\ref{tab:ablation_results} show that contextual filtering substantially outperforms summarization across all metrics. This suggests that privacy reasoning benefits from selective retrieval: filtering preserves scenario-relevant beliefs and supports more accurate, human-like concern prediction than coarse summarization.

\begin{table*}[t]
\centering
\small
\renewcommand{\arraystretch}{1.2}

\begin{minipage}{\textwidth}
\captionof{table}{Main results on privacy-concern prediction. 
We compare (1) Privacy Persona, (2) RAG, (3) naive baseline, and (4) our full \(\PRA\) model. 
Performance is measured using \textbf{Accuracy}, \textbf{Recall}, and \textbf{F1-score}, 
as evaluated by the privacy-concern classifier \Judge. 
Results are averaged over three independent runs, with standard deviations reported.}
\label{tab:main_results}

\begin{tabularx}{\textwidth}{@{}l *{3}{>{\centering\arraybackslash}X}@{}}
\toprule
\textbf{Agent Structure} & \textbf{Accuracy \(\uparrow\)} & \textbf{Recall \(\uparrow\)} & \textbf{F1-score \(\uparrow\)} \\
\midrule
Naive Baseline & \(0.747 \pm 0.014\) & \(0.319 \pm 0.017\) & \(0.335 \pm 0.018\) \\
Privacy Persona & \(0.779 \pm 0.013\) & \(0.401 \pm 0.010\) & \(0.403 \pm 0.011\) \\
RAG & \(0.803 \pm 0.005\) & \(0.498 \pm 0.016\) & \(0.449 \pm 0.010\) \\
\(\PRA\) & \(\textbf{0.829} \pm 0.011\) & \(\textbf{0.533} \pm 0.019\) & \(\textbf{0.473} \pm 0.015\) \\
\bottomrule
\end{tabularx}
\end{minipage}

\vspace{0.6em}

\begin{minipage}{\textwidth}
\captionof{table}{Ablation study on privacy-concern prediction. 
We compare (1) a naive baseline, (2) unstructured atomic distillation, 
(3) summary-based contextual filtering, and (4) our full \(\PRA\) model. 
Performance is measured using \textbf{Accuracy}, \textbf{Recall}, and \textbf{F1-score}, 
as evaluated by the privacy-concern classifier \Judge. 
Results are averaged over three independent runs, with standard deviations reported.}
\label{tab:ablation_results}

\begin{tabularx}{\textwidth}{@{}l *{3}{>{\centering\arraybackslash}X}@{}}
\toprule
\textbf{Agent Structure} & \textbf{Accuracy \(\uparrow\)} & \textbf{Recall \(\uparrow\)} & \textbf{F1-score \(\uparrow\)} \\
\midrule
Naive Baseline & \(0.747 \pm 0.014\) & \(0.319 \pm 0.017\) & \(0.335 \pm 0.018\) \\
Plain Distillation & \(0.801 \pm 0.003\) & \(0.472 \pm 0.008\) & \(0.435 \pm 0.019\) \\
Summary & \(0.782 \pm 0.009\) & \(0.410 \pm 0.012\) & \(0.433 \pm 0.016\) \\
\(\PRA\) & \(\textbf{0.829} \pm 0.011\) & \(\textbf{0.533} \pm 0.019\) & \(\textbf{0.473} \pm 0.015\) \\
\bottomrule
\end{tabularx}
\end{minipage}
\end{table*}
\begin{figure}
    \centering
    \includegraphics[width=\linewidth]{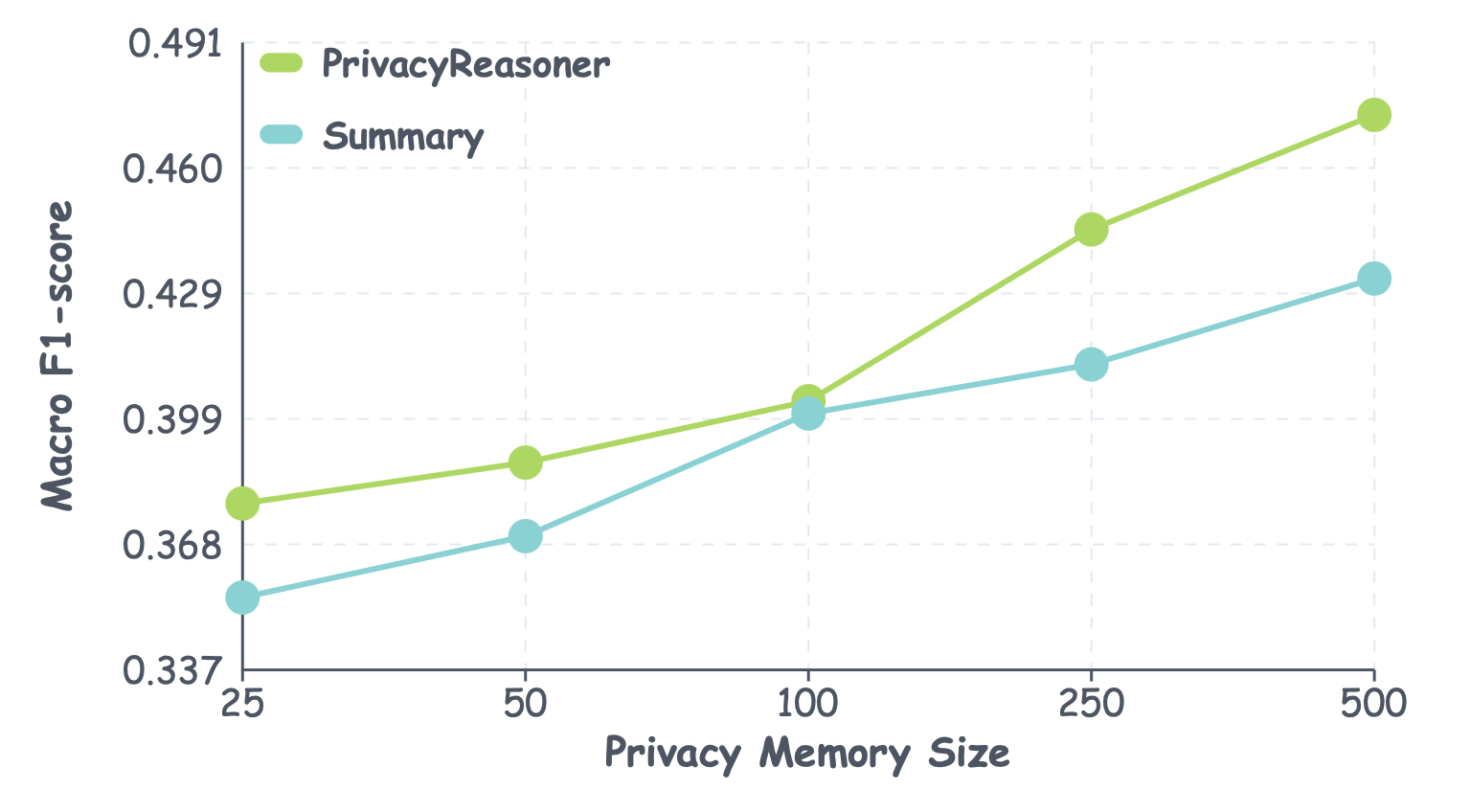}
\caption{Effect of privacy memory size on privacy concern prediction. Increasing the number of historical user comments available to both \(\PRA\) and \textbf{summary-based} variant. We report \textbf{macro F1-scores} as the evaluation metric, as well as the standard deviation.}
    \label{fig:memory}
\end{figure}

\begin{table}[t]
\centering
\small
\caption{Domain transfer performance of the \(\PRA\) agent. The agent is constructed using user comment histories exclusively from the AI domain and evaluated on privacy-sensitive discussions in e-commerce and healthcare. }

\label{tab:domain_transfer}
\renewcommand{\arraystretch}{1.2}
\begin{tabularx}{\columnwidth}{l@{\hspace{4em}}c@{\hspace{3em}}c}
\toprule
\textbf{Metric} & E-commerce & Health-care \\
\midrule
\textbf{Accuracy \(\uparrow\)}  & \(0.807 \pm 0.007\) & \(0.802 \pm 0.011\) \\
\textbf{Recall \(\uparrow\)}    & \(0.430 \pm 0.015\) & \(0.428 \pm 0.018\) \\
\textbf{F1-score \(\uparrow\)} & \(0.410 \pm 0.014\) & \(0.407 \pm 0.012\) \\
\bottomrule
\end{tabularx}
\end{table}

\begin{table*}[t]
\centering
\small
\caption{Results on privacy-concern prediction with human interview data. 
Performance is measured using \textbf{Accuracy}, \textbf{Recall}, and \textbf{F1-score}, 
as evaluated by the privacy-concern classifier \Judge. 
Results are averaged over three independent runs, with standard deviations reported.}
\label{tab:main_results_human}

\renewcommand{\arraystretch}{1.2}
\begin{tabularx}{\textwidth}{@{}l *{3}{>{\centering\arraybackslash}X}@{}}
\toprule
\textbf{Agent Structure} & \textbf{Accuracy \(\uparrow\)} & \textbf{Recall \(\uparrow\)} & \textbf{F1-score \(\uparrow\)} \\
\midrule
Naive Baseline & \(0.758 \pm 0.033\) & \(0.508 \pm 0.021\) & \(0.413 \pm 0.029\) \\
Privacy Persona & \(0.767 \pm 0.031\) & \(0.502 \pm 0.038\) & \(0.409 \pm 0.029\) \\
RAG & \(0.803 \pm 0.015\) & \(0.539 \pm 0.018\) & \(0.435 \pm 0.033\) \\
\(\PRA\) & \(\textbf{0.810} \pm 0.032\) & \(\textbf{0.570} \pm 0.013\) & \(\textbf{0.469} \pm 0.012\) \\
\bottomrule
\end{tabularx}
\end{table*}

\paragraph{Privacy Memory Size.}~\label{sec:privacymemory}
We analyze the effect of privacy memory size by varying the number of historical user comments from 25 to 500 and measuring concern faithfulness. As shown in Figure~\ref{fig:memory}, performance improves monotonically with memory size, indicating that richer user history better supports reconstruction of user-specific privacy reasoning. Larger memory enables the agent to capture more nuanced belief structures and patterns. However, gains diminish beyond a threshold for both agent variants, suggesting decreasing marginal utility as cognitive filters saturate. This trend is consistent with bounded rationality, where decision-making relies on a limited subset of salient information rather than exhaustive recall.

\subsection{Generalizability}

\begin{table*}[t]
\centering
\small
\caption{Concern prediction performance on the \textit{Apple CSAM} discussion selected from HackerNews Dataset. Metrics are computed at the individual user level and subsequently aggregated across all users participating in the post. Results are averaged over three independent runs, with standard deviations reported.}

\label{tab:csam_results}
\renewcommand{\arraystretch}{1.2}
\begin{tabularx}{\textwidth}{@{}l *{3}{>{\centering\arraybackslash}X}@{}}
\toprule
\textbf{Agent Structure} & \textbf{Accuracy \(\uparrow\)} & \textbf{Recall \(\uparrow\)} & \textbf{F1-score \(\uparrow\)} \\
\midrule
Naive Baseline & \(0.795 \pm 0.011\) & \(0.367 \pm 0.009\) & \(0.330 \pm 0.012\) \\
Privacy Persona & \(0.801 \pm 0.016\) & \(0.345 \pm 0.017\) & \(0.339 \pm 0.018\) \\
RAG & \(0.810 \pm 0.017\) & \(0.477 \pm 0.013\) & \(0.435 \pm 0.021\) \\
\(\PRA\)  & \(\textbf{0.821} \pm 0.015\) & \(\textbf{0.508} \pm 0.017\) & \(\textbf{0.453} \pm 0.014\) \\
\bottomrule
\end{tabularx}
\end{table*}

\subsubsection{Domain Transfer} 
Domain transfer evaluates an agent’s ability to generalize beyond the domain in which its user profile was constructed. We test whether an agent built from artificial intelligence discussions can accurately simulate privacy concerns in different contexts. Concretely, we construct user profiles solely from AI-related posts on \emph{Hacker News} and evaluate the resulting agent in other domains like e-commerce and healthcare. This setting is challenging: different domains may have different privacy focus. For both domains, we sample \(50\) comments for evaluation.
\paragraph{Results and Analysis.}  
Table~\ref{tab:domain_transfer} shows that \(\PRA\), even when constructed solely from AI-related user histories, retains strong performance on e-commerce and healthcare discussions. This suggests that its cognitive and contextual mechanisms capture domain-agnostic privacy reasoning patterns rather than overfitting to AI-specific content. Overall, these results demonstrate that \(\PRA\) learns transferable cognitive structures that support effective cross-domain reasoning even from limited, domain-specific histories.

\subsubsection{User Transfer} 
Users vary widely in the amount of historical data available. For those with rich comment histories, constructing a detailed privacy memory is relatively straightforward, enabling the agent to model preferences and concerns with greater fidelity. Conversely, sparse histories pose a greater challenge, as limited evidence constrains the agent’s ability to reconstruct nuanced belief structures and reasoning patterns. To evaluate the robustness of our approach under different data conditions, we examine how \(\PRA\) performs across varying levels of historical information. Unlike the post-oriented sampling used in~\Cref{sec:privacymemory}, where each user contributes only a single comment, here we sample \(10\) users with varying amounts of comment history and evaluate the model on \(15\) sampled comments per user to assess how performance scales with the richness of user data.

\begin{figure}
    \centering
    \includegraphics[width=\linewidth]{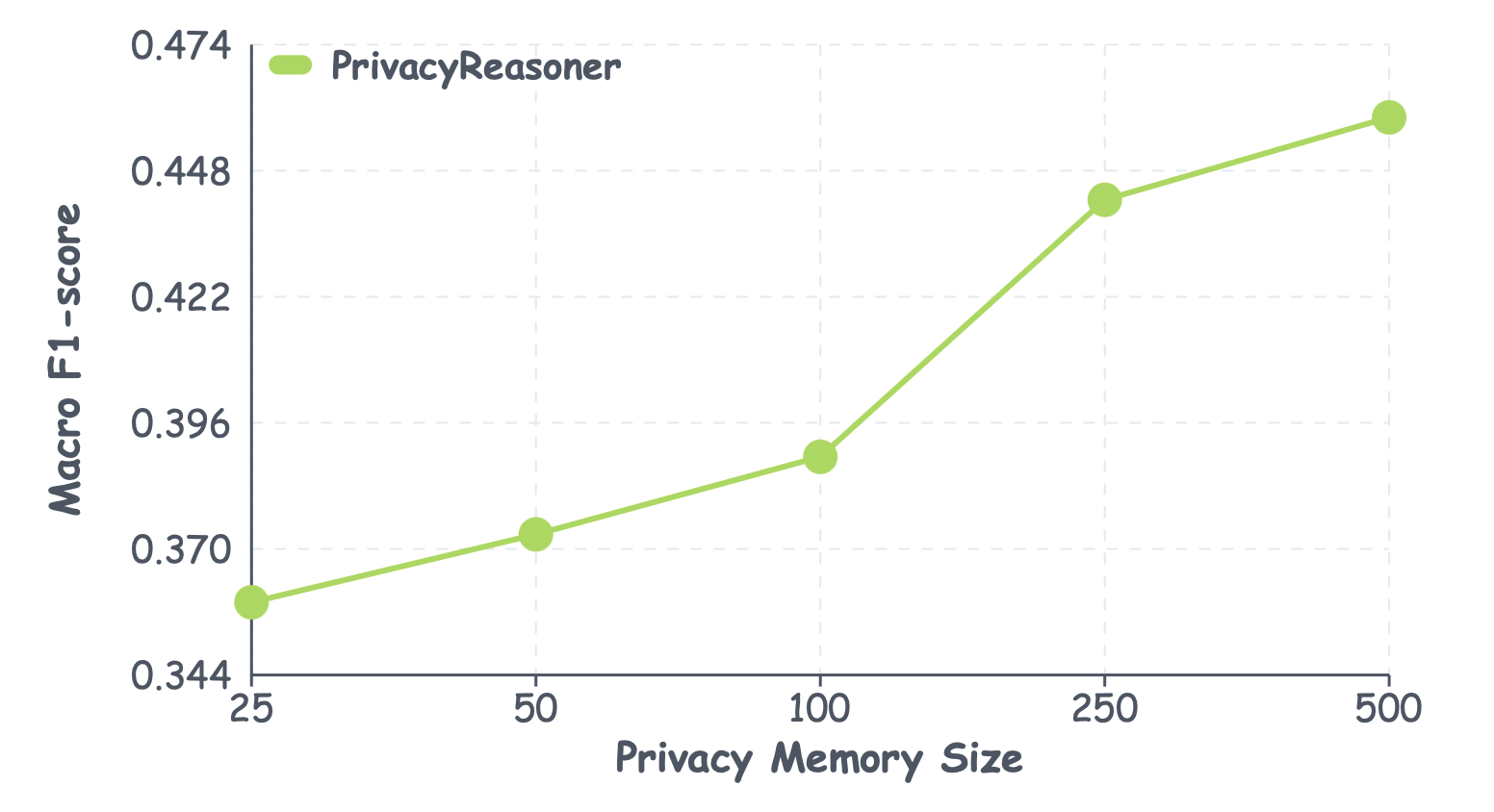}
\caption{User transfer performance of the  \(\PRA\) agent constructed with varying amounts of user comment history.}
    \label{fig:usertransfer}
\end{figure}

\paragraph{Results and Analysis.}  
Figure~\ref{fig:usertransfer} shows that the performance of \(\PRA\) decreases as the amount of available user history diminishes. This is expected, as sparser histories make it more challenging to faithfully reconstruct a user’s underlying privacy reasoning due to limited contextual evidence. Nevertheless, \(\PRA\) still achieves a non-trivial macro F1-score of around \(0.4\) in low-data settings, demonstrating its ability to generalize and simulate user-specific reasoning even when historical information is scarce.

\subsection{Generalization across data sets}
In addition to \emph{Hacker News}, we conduct experiments on the real human interview dataset from~\citep {10.1145/3613904.3642500}, which captures more structured and reflective expressions of privacy concerns. The dataset contains 1306 comments. We construct the evaluation by reserving \(15\) comments as few-shot examples for \(\Judge\) and splitting the remaining data into \(70\%\) for training and \(30\%\) for testing. This setup introduces distributional shifts in both language style and content, providing a more stringent test of generalization.

\paragraph{Results and Analysis.}  
As shown in~\Cref{tab:main_results_human}, results are highly consistent with those on \emph{Hacker News}. Despite differences in data source and linguistic characteristics, \(\PRA\) maintains comparable performance, suggesting that it captures underlying privacy reasoning patterns rather than overfitting to dataset-specific artifacts. Overall, these findings demonstrate the robustness of \(\PRA\) across heterogeneous data settings.

\section{Case Study: Apple CSAM Controversy}
Finally, as a case study, we revisit the Apple CSAM incident introduced at the beginning of this paper and apply our \(\PRA\) framework to predict how individual users might respond to such a privacy-sensitive event. We identify the relevant discussion post on \emph{Hacker News} and deploy different agent structures mentioned above to simulate and analyze the range of privacy concerns emerging in this context. Specifically, the post is titled ``Apple enabling client-side CSAM'', posted on August 2021. We sample 75 comments from the post, involving 75 distinct users.

\paragraph{Results and Analysis.}  
Table~\ref{tab:csam_results} presents the concern prediction results across four agent variants. Overall, \(\PRA\) achieves the strongest performance across all metrics, demonstrating the effectiveness of selective belief activation and context-sensitive reasoning. These results indicate that \(\PRA\) can provide non-trivial and actionable insights into how individuals are likely to respond to the Apple CSAM incident, thereby equipping decision-makers with a more realistic understanding of public privacy concerns.

\section{Conclusion}
We introduced \(\PRA\), a privacy reasoning agent that models individual-level privacy concerns by constructing structured privacy memories and leveraging LLMs to simulate personalized decision-making. Across user, domain, and dataset transfer settings, \(\PRA\) consistently outperforms baselines, demonstrating robustness to data sparsity and strong generalization across contexts. These results suggest that \(\PRA\) captures underlying, domain-agnostic privacy reasoning patterns rather than surface-level signals. More broadly, our findings highlight the potential of LLM-based agents as scalable proxies for human privacy reasoning, enabling proactive evaluation of privacy-sensitive systems and reducing reliance on costly human studies.

\section{Limitations}
While \(\PRA\) demonstrates promising results in modeling individualized privacy reasoning, several limitations warrant discussion. First, our evaluation relies on publicly available Hacker News discussions, which, despite their richness, represent a specific demographic of technically oriented users and may not generalize to broader populations with different privacy attitudes and discourse styles. Second, the LLM-as-a-Judge evaluation framework, while carefully calibrated and aligned with human annotation, is itself subject to potential biases and misclassifications inherent to large language models. Moreover, our study primarily focuses on textual reasoning and does not capture multimodal signals (e.g., behavioral or interactional data) that may shape privacy attitudes in real-world settings. Finally, while the framework shows cross-domain transferability, further research is needed to assess its robustness in more dynamic contexts, evolving privacy norms, and under adversarial or strategic user behaviors. Third, this work only focuses on one-level response, further work could explore scenarios like privacy debates or conversations to yield richer information. 
\bibliography{custom}
\newpage
\appendix
\section*{Agent Specification}

All agent components in \textsc{PrivacyReasoner} are implemented using \texttt{gpt-4o-mini}~\cite{openai2024gpt4ocard} as the underlying language model except for \textsc{PrivacyConcernJudge} which uses \texttt{gpt-4o}~\cite{openai2024gpt4ocard} . Unless otherwise noted, we set the decoding temperature to \(0.0\) across all generation stages, including belief extraction, context filtering, persona classification, and synthetic comment generation. This low temperature reduces randomness and improves faithfulness in reasoning tasks. The same model and temperature setting are also used for the \textsc{PrivacyConcernJudge} to ensure consistency in evaluation. 

\section*{Potential Risk}
 While PrivacyReasoner advances the modeling of individualized privacy reasoning, its deployment also raises important risks. First, though we used publicly available dataset where all users are anonymous, constructing user-specific privacy minds from historical comments on other corpus may introduce the possibility of privacy violations or misuse, especially if sensitive data is collected, stored, or analyzed without sufficient safeguards. Second, the reliance on large language models as both generator and evaluator may amplify underlying model biases or produce false inferences about user beliefs, potentially leading to skewed simulations and inaccurate representations of individuals. Careful governance, transparency, and ethical safeguards are therefore essential to ensure that PrivacyReasoner’s capabilities are used responsibly.

\section*{Privacy Concern Judge Specification}
To evaluate the faithfulness of generated comments with respect to privacy reasoning, we implement a \textsc{PrivacyConcernJudge}, an LLM-as-a-judge classifier based on \texttt{gpt-4o} (OpenAI, 2024). The judge models concern detection as a 14-way multi-label classification task and operates with a decoding temperature of \(0.0\) to ensure deterministic and faithful judgments. It is prompted in a few-shot fashion with representative examples of each concern category to align its classification behavior with human annotations.

\subsection{Guidelines and Edge Cases}

\begin{itemize}
    \item Multiple concerns may co-occur in a single comment and should all be labeled.
    \item General policy or technical critiques should not be labeled unless explicitly connected to privacy harms, norms, or user rights.
    \item Annotators should avoid inferring intent beyond what is reasonably supported by the text.
    \item Ambiguous cases should be resolved conservatively, favoring precision over recall.
\end{itemize}

\subsection{Intended Use}

This annotation codebook supports (1) human labeling of privacy concerns in online discourse, (2) few-shot prompting and calibration of LLM-based privacy concern judges, and (3) reproducible evaluation of synthetic comment faithfulness in individualized privacy reasoning tasks.

\paragraph{Concern Taxonomy.}  
We follow the 14 privacy concern categories introduced by \citet{10.1145/3613904.3642500}, the detailed few-shot examples are listed in~\Cref{tab:privacy_judge_examples}.

\begin{table*}[h]
\centering
\small
\renewcommand{\arraystretch}{1.3}
\begin{tabularx}{\textwidth}{lX}
\toprule
\textbf{Privacy Concern} & \textbf{Representative Example} \\
\midrule
Lack of trust for algorithms & I feel extremely uncomfortable because I don't want a company “re-shaping” my shopping habits. That is ridiculous. Also, I would not want to be inundated with offers based on a company's “prediction” that might not be correct. \\
Lack of an alternative choice & I feel extremely uncomfortable because if I am not interested in the social network service, I have no choice but to let them share my email with the social network team and I have no idea what they will do with that data. \\
Insufficient anonymization & I feel somewhat uncomfortable with this because for me progress pictures are private and something I only want access to. I wouldn't want them brought up on a Google search for sure because some of them might be potentially embarrassing. \\
Lack of respect for autonomy & I feel extremely uncomfortable because I feel like the company is delving too much into my personal life by developing an algorithm. I don't like the idea of a company deciding what content is relevant and engaging to me. \\
Bias or discrimination & I feel uncomfortable because this is unethical all the way around. A person should not be charged based on their phone's battery. \\
Data leakage & I feel uncomfortable for so many reasons! The biggest reason is the fact that it collects credit card numbers and other PII. I would be concerned that the information might be hacked and someone would get my PII. \\
Lack of informed consent & I feel uncomfortable because they should not be recording other people who have not opted in to this service, as they did not ask permission to do this. \\
Invasive monitoring & I would feel somewhat uncomfortable using this tracking app; I would feel it could be used by my employer to invade my privacy. \\
Data commodification & I feel uncomfortable with this because I see no reason whatsoever that a retail store would need to sell data to an insurance company. \\
No control & I feel uncomfortable with my data being linked to my credit card. They could turn off my card if you don't follow the PC thinking of the day. \\
High risks & I feel uncomfortable as your safety could be at risk. If you're somewhere and your battery is low and you can't or don't want to pay the fee you're stuck there. \\
Unexpectation & Weight loss and food are not super personal. However, it is not something I'd want my friends and family or even strangers to see on Google. This is especially true with before pictures. \\
Lack of protection for the vulnerable & I feel extremely uncomfortable and angry about this because in these times since the strikedown of Roe, this employer is essentially offering a \$1 gift card a day for information that if exposed, could lead to a woman's criminalization, jail time and even execution simply for exercising what should be her right to her own bodily autonomy. \\
\bottomrule
\end{tabularx}
\caption{Few-shot examples used to guide the \textsc{PrivacyConcernJudge} for each privacy concern category, all data are interviewee data selected from~\citet{10.1145/3613904.3642500}.}
\label{tab:privacy_judge_examples}
\end{table*}

\section*{Prompt Templates}

We include the core prompt templates used in \textsc{PrivacyReasoner} for key tasks, enabling reproducibility and providing transparency into how the agent components interact with the underlying language model. The detailed specification such as specific user privacy memory, post context, few shot examples for privacy judge are omitted.

\paragraph{Privacy Distillation.}  
To build a user’s privacy memory, we extract atomic beliefs and reasoning patterns from their historical comments.

\begin{quote}
\small
\texttt{You are an expert in privacy psychology analyzing an individual's historical privacy-related texts. Following the APCO (Antecedents Privacy Concerns Outcomes) framework, extract atomic statements about the individual's privacy antecedents along five dimensions: [...]. For each dimension, extract distinct atomic statements that capture specific tendencies evidenced in their past expressions.}
\end{quote}

\paragraph{Contextual Filtering.}  
At inference time, we activate the most contextually relevant beliefs from the privacy memory to simulate bounded working memory.

\begin{quote}
\small
\texttt{Given the following user beliefs and the post context, select the most relevant beliefs that are likely to be activated in this situation. Re-rank them by their relevance and confidence for generating a response.}
\end{quote}

\paragraph{Synthetic Comment Generation.}  
Finally, the agent generates a synthetic comment simulating how the user would respond to a new privacy-related event.

\begin{quote}
\small
\texttt{You are [user id] when it comes to privacy. Given the following discussion post and the selected beliefs from your privacy memory, write a comment that reflects the user's likely privacy concerns, reasoning, and tone. The comment should be natural, context-aware, and faithful to the user's historical reasoning style. Favor details to generality.}
\end{quote}

\paragraph{Privacy Concern Judgment.}  
The \textsc{PrivacyConcernJudge} is prompted with few-shot examples (see Table~\ref{tab:privacy_judge_examples}) to classify expressed concerns.

\begin{quote}
\small
\texttt{Given the following comment, identify which of the 14 privacy concerns it expresses. Return a JSON object with concern names as keys and 1/0 values indicating presence or absence. Base your decision strictly on explicit evidence in the text.}
\end{quote}

\section*{Qualitative Examples}~\label{sec:qualitative}

We examine several representative examples where \(\PRA\) captures nuanced privacy reasoning that the baseline approach misses. Table~\ref{tab:qualitative} compares actual user comments with synthetic comments generated by \(\PRA\) and a baseline model across three privacy-sensitive contexts: IoT cybersecurity regulation, Triplebyte’s default public profile controversy, and Apple’s proposed CSAM client-side scanning feature.

\begin{table*}[t]
\centering
\small
\caption{Qualitative comparison of actual comments, \(\PRA\)-generated comments, and baseline-generated comments across three representative discussion threads.}
\label{tab:qualitative}
\renewcommand{\arraystretch}{1.2}
\begin{tabularx}{\textwidth}{p{2.1cm} p{3.2cm} p{4.5cm} p{4.5cm}}
\toprule
\textbf{Thread Context} & \textbf{Actual Comment} & \textbf{\(\PRA\) Response} & \textbf{Baseline Response} \\
\midrule
\textbf{FCC IoT Labeling Program} & 
``What makes IoT devices special, and warrants carve outs for security... I can’t decide if I’m more concerned about my Wyze camera sending data than my water heater leaking its temperature.'' &
``As someone who values data privacy and security, I appreciate Commissioner Simington's efforts to address the vulnerabilities in IoT devices... I urge consumers to voice concerns about data privacy and potential data-sharing between companies...'' &
``As a consumer, I appreciate Commissioner Simington's efforts... However, I believe privacy concerns should also be a key consideration... How will the FCC ensure that manufacturers are not compromising user privacy...'' \\
\midrule
\textbf{Triplebyte Privacy Backlash} &
``Rather than safeguarding the fact that you are or were job searching, we threatened exposure... This was a breach of trust...'' &
``As someone who values privacy and security, I appreciate Ammon’s sincere apology and recognition of the breach of trust... It’s concerning how easily personal information can be mishandled, especially in a job search context where confidentiality is crucial...'' &
``I appreciate Triplebyte's swift response... Companies must prioritize user privacy... I hope this serves as a learning experience...'' \\
\midrule
\textbf{Apple CSAM Proposal} &
``It’s quite easy to extrapolate this... First it’s iPhone photos, then all iCloud files, then local Mac files... What’s the utopian perspective that balances this risk?'' &
``While the intention to protect children is noble, the potential for misuse and the slippery slope of expanding this technology is a valid concern... Apple must remain transparent about algorithms and safeguards to prevent unauthorized access or broader surveillance...'' &
``I appreciate Apple's efforts to implement client-side scanning... However, I have some concerns about how this may impact my data security... I hope Apple will be transparent and strike a balance between protection and rights...'' \\
\bottomrule
\end{tabularx}
\end{table*}

\paragraph{Analysis.}  
Across all three scenarios, \(\PRA\) consistently captures finer-grained reasoning patterns that are present in real user discourse but absent from baseline outputs. For instance, in the IoT labeling case, \(\PRA\) highlights downstream risks such as \emph{inter-company data sharing} and explicitly calls for \emph{user participation in shaping regulatory standards}, closely mirroring human concerns about autonomy and accountability. In the Triplebyte example, \(\PRA\) situates the apology within the broader context of \emph{job-search confidentiality}, emphasizing the specific stakes of privacy violation rather than generic appeals to “learning from mistakes” as the baseline does. Finally, in the Apple CSAM case, \(\PRA\) identifies the \emph{slippery slope toward expanded surveillance} and calls for \emph{algorithmic transparency} and \emph{false-positive safeguards}, while the baseline remains surface-level and avoids articulating the deeper structural risks raised by users. These examples illustrate how \(\PRA\)’s cognitively grounded reasoning enables it to surface implicit concerns and contextual nuances that baseline approaches systematically overlook.

\section{Additional Experiment Results}
\subsection{Different Base Model}\label{sec:base_model}
\begin{table*}[t]
\centering
\small
\caption{Results on privacy-concern prediction with \texttt{gemini-3-flash-preview}. 
We compare (1) Privacy Persona, (2) RAG, (3) naive baseline, and (4) our full \(\PRA\) model. 
Performance is measured using \textbf{Accuracy}, \textbf{Recall}, and \textbf{F1-score}, 
as evaluated by the privacy-concern classifier \Judge. 
Results are averaged over three independent runs, with standard deviations reported.}
\label{tab:main_results_gemini}

\begin{tabular}{lccc}
\toprule
\textbf{Agent Structure} & \textbf{Accuracy \(\uparrow\)} & \textbf{Recall \(\uparrow\)} & \textbf{F1-score \(\uparrow\)} \\
\midrule
Naive Baseline & \(0.785 \pm 0.068\) & \(0.491 \pm 0.005\) & \(0.389 \pm 0.019\) \\
Privacy Persona & \(0.792 \pm 0.027\) & \(0.505 \pm 0.039\) & \(0.412 \pm 0.032\) \\
RAG & \(0.798 \pm 0.058\) & \(0.534 \pm 0.043\) & \(0.426 \pm 0.055\) \\
\(\PRA\) & \(\textbf{0.801} \pm 0.054\) & \(\textbf{0.587} \pm 0.022\) & \(\textbf{0.467} \pm 0.067\) \\
\bottomrule
\end{tabular}
\end{table*}

As for potential biases inherent in the GPT-4o process, we additionally conduct experiments using \textbf{gemini-3-flash-preview}. As shown in~\Cref{tab:main_results_gemini}, the results are highly consistent with those obtained using \textbf{GPT-4o}, suggesting that our findings are robust across different LLM backbones and are not driven by model-specific biases.

\section*{AI Tool Use}
GPT-5~\citep{openai2025gpt5} was employed in this project to support both manuscript preparation and code development, serving as an aid for drafting, refinement, and implementation tasks.

\end{document}